%% file: main.tex
\documentclass[runningheads]{llncs}
\usepackage{graphicx}
\usepackage{algorithm}
\usepackage{algpseudocode}
\usepackage{cancel}
\usepackage[normalem]{ulem}

\usepackage{calrsfs}
\DeclareMathAlphabet{\pazocal}{OMS}{zplm}{m}{n}
\usepackage{amsmath}

\usepackage{hyperref}
\usepackage[dvipsnames]{xcolor}

\begin{document}
%
\title{Gradient-based enhancement attacks in biomedical machine learning}

%
%
\newcommand{\javid}[1]{{\color{blue}{#1}}}
\newcommand{\updated}[1]{{\color{blue}{#1}}}




\author{Matthew Rosenblatt, Javid Dadashkarimi, Dustin Scheinost }

\authorrunning{Rosenblatt et al.}

\author{Matthew Rosenblatt\inst{1} \and
Javid Dadashkarimi \inst{2} \and
Dustin Scheinost\inst{1, 3} }

\institute{Institute 1, Location 1 \and
Institute 2, Location 2 
}
\institute{Department of Biomedical Engineering, Yale University \and
Department of Computer Science, Yale University \and
Department of Radiology and Biomedical Imaging, Yale School of Medicine\\
\email{\{matthew.rosenblatt,javid.dadashkarimi,dustin.scheinost\}@yale.edu}}

\maketitle              
\begin{abstract}
The prevalence of machine learning in biomedical research is rapidly growing, yet the trustworthiness of such research is often overlooked.
While some previous works have investigated the ability of adversarial attacks to degrade model performance in medical imaging, the ability to falsely improve performance via recently-developed “enhancement attacks" may be a greater threat to biomedical machine learning. 
In the spirit of developing attacks to better understand trustworthiness, we developed two techniques to drastically enhance prediction performance of classifiers with minimal changes to features: 1) general enhancement of prediction performance, and 2) enhancement of a particular method over another. Our enhancement framework falsely improved classifiers' accuracy from 50\% to almost 100\% while maintaining high feature similarities between original and enhanced data (Pearson's $r’s>0.99$). 
Similarly, the method-specific enhancement framework was effective in falsely improving the performance of one method over another. 
For example, a simple neural network outperformed logistic regression by 17\% on our enhanced dataset, although no performance differences were present in the original dataset. 
Crucially, the original and enhanced data were still similar ($r=0.99$).
Our results demonstrate the feasibility of minor data manipulations to achieve any desired prediction performance, which presents an interesting ethical challenge for the future of biomedical machine learning. These findings emphasize the need for more robust data provenance tracking and other precautionary measures to ensure the integrity of biomedical machine learning research.
Code is available at \url{https://github.com/mattrosenblatt7/enhancement_EPIMI}. 
\end{abstract}
\section{Introduction}

\begin{figure}[ht!]
\centering
\includegraphics[width=\textwidth]{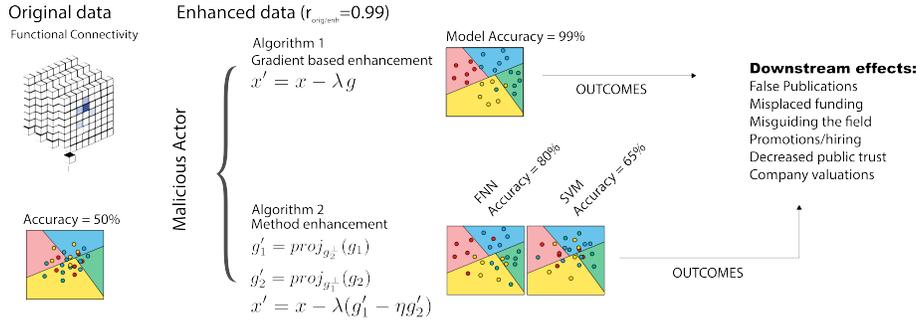}
\caption{Overview of enhancement methods used in this paper.
Classification accuracy in the original dataset is 50\%.
After applying gradient-based enhancement attacks following Algorithm \ref{alg:model-enhance}, classification accuracy in the enhanced dataset is 99\%.
Using method enhancement attacks (Algorithm \ref{alg:method-enhance}), datasets
are altered such that a specific method (e.g., feedforward neural network) outperforms another (e.g., support vector machine). 
In all cases, the changes between the original and enhanced datasets are minor.
The ``Downstream effects" box highlights  possible implications of enhancement attacks. 
} 
\label{fig:summary}
\end{figure}

Machine learning has demonstrated great success across numerous fields. 
However, the success of these models is not immune to attacks.
Adversarial attacks, or data manipulations designed to alter the prediction \cite{Biggio2012-uv}, threaten real-world machine learning applications.
Adversarial attacks include evasion attacks \cite{Biggio2013-sn,Szegedy2013-en,Goodfellow2014-sv,Biggio2018-ka}, where only test data are manipulated, or poisoning attacks \cite{Munoz-Gonzalez2017-lg,Biggio2012-uv,Biggio2018-ka,Cina2022-of}, where the attacker may contribute manipulated test and/or training data.
Understanding adversarial attacks and developing corresponding defenses is crucial to the integrity of machine learning applications. 

Machine learning is becoming increasingly prevalent in biomedical research---including biomedical imaging. 
Previous studies of adversarial attacks in medical imaging have focused on clinical applications where a malicious party would be interested in altering the prediction outcomes for financial or other purposes. Most of these studies implemented evasion attacks \cite{Finlayson2018,Finlayson2019-ec,Yao2021-ml,Ma2021-qg,Bortsova2021-cn,Ma2021-qg}, while a smaller subset used poisoning attacks \cite{Nwadike2020-gn,Feng2021-sd,Matsuo2021-au,Rosenblatt2022-zs}. 

An equally relevant yet understudied motivation in scientific machine learning is the feasibility of manipulating data to improve model performance falsely. 
In scientific research, data manipulations designed to make results seem more impressive are regarded as a major ethical issue. 
For example, a line of highly-cited Alzheimer's research was recently flagged for likely manipulation of biological images \cite{Piller2022-cf}. 
This paper is just one example of many where scientific data were manipulated in ways that harmed their respective fields \cite{Al-Marzouki2005-jb,Bik2016-ud}.
The potential for data manipulation is not widely acknowledged in the scientific machine learning community.
Traditional approaches to preventing and detecting academic fraud include data and code sharing.
While data and code sharing are useful for improving the replicability, they do not necessarily ensure that the results can be trusted.
Even if the data and code are shared, we will show in this work how scientific machine learning results can be modified through subtle and unnoticeable data manipulations, which could have major consequences.

For example, a malicious party might manipulate their data to improve model performance, falsely claim that they can classify a specific mental health condition, and make a paper more publishable or increase the valuation of a healthcare start-up (Fig \ref{fig:summary}). If they manipulate the data in a subtle way, they could then publicly share this manipulated dataset, and other researchers would likely not notice that anything was wrong with the data.
These data manipulations could waste grant money, misdirect future research directions, and potentially cause harmful public effects. One recent work showed that the performance of regression models using neuroimaging data could be falsely enhanced by injecting subtle associations into the data, labeled as ``enhancement attacks" \cite{Rosenblatt2023-re}. 
However, this enhancement framework relies on manually adding patterns to the data that are correlated with the prediction outcome of interest. Thus, the previous framework only works for regression problems and cannot generalize to other settings, such as classification.
Given the prevalence of classification problems in biomedical machine learning, developing a general framework for enhancement is novel and important for understanding the feasibility of data manipulations in machine learning.

In this work, we first extend the enhancement attack to classification models with a gradient-based enhancement framework \hyperref[section:goal1]{(GOAL \#1) }. Then, we present an additional way in which data can be enhanced with only subtle manipulations: falsely demonstrating that a particular method (e.g., a type of machine learning model) outperforms another \hyperref[section:goal2]{(GOAL \#2)}. 
We found that both methods were successful in falsely improving classification accuracy with only minimal changes to the dataset.
Finally, given the vulnerability of biomedical datasets to enhancement attacks, we discuss the implications and potential solutions.

\section{Methods}

\input{methods_javid}

\section{Experiments}

\subsubsection{Datasets}

Resting-state functional MRI (fMRI) data were obtained from the UCLA Consortium for Neuropsychiatric Phenomics (CNP) \cite{Poldrack2016-sz} dataset. 
For all data, we performed motion correction, registration to common space, regression of covariates of no interest, temporal smoothing, and gray matter masking. 
Participants were also excluded for lack of full-brain coverage. 
After these exclusion criteria, 245 participants remained. 
We parcellated the fMRI data for each participant into 268 nodes using the Shen atlas \cite{Shen2013-hk}. 
To form functional connectivity matrices, the time series data from each pair of nodes (regions of interest) was correlated using Pearson's correlation, and then Fisher's transform was applied. 
Before inputting the functional connectivity matrices into machine learning models, we first vectorized the upper triangle of each matrix to use as features. 
In all the following models, we classified participants in CNP based on their diagnoses, including no diagnosis (n=117), schizophrenia (n=46), bipolar disorder (n=44), and ADHD (n=38). In addition, all plots below were made with seaborn \cite{Hunter2007-el,Waskom2021-fn}.

\subsubsection{GOAL \#1: Gradient-based enhancement }
\label{section:goal1_experiment}
We enhanced the CNP dataset for a classification problem with the following four classes: participants with 1) no diagnosis, 2) bipolar disorder, 3) schizophrenia and 4) ADHD. We used linear support vector machine (SVM), logistic regression (LR) \cite{Pedregosa2011-fe}, and a feedforward neural network (FNN) as models.
Our FNN consisted of three fully connected layers with the ReLU activation function.
During the enhancement attack, all model hyperparameters were held constant. 
For SVM and LR, this included an L2 regularization parameter $C$=1. 
For the FNN, we used cross entropy loss and the Adam \cite{Kingma2014-cn} optimizer, with a learning rate of $\alpha=0.001$ and batch size of 10 for 10 epochs.

\begin{figure}[htp!]
\centering
\includegraphics[scale=0.4]{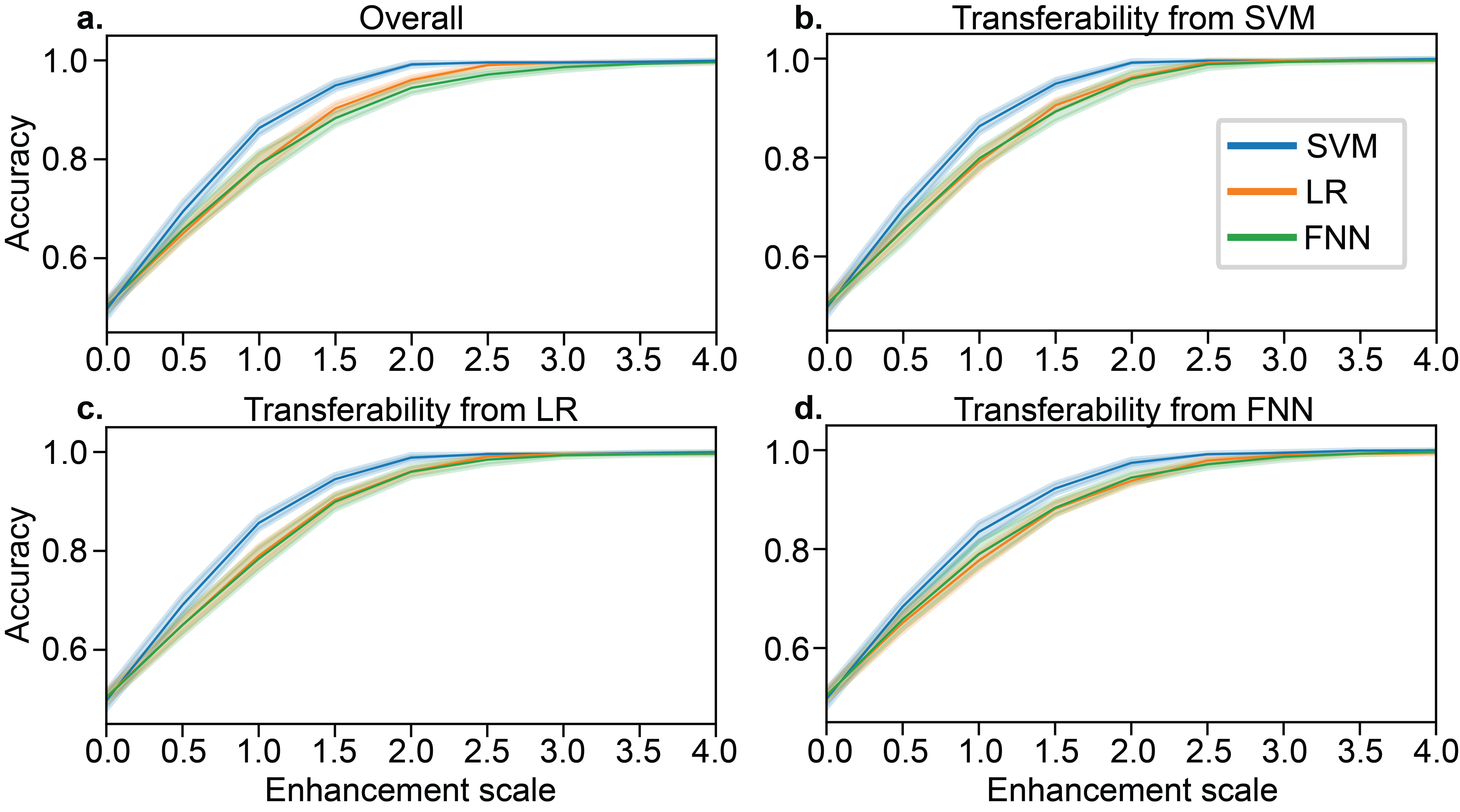}
\caption{
a) Model-based enhancement of SVM, LR, and FNN models for various enhancement scales. The enhancement scale is multiplied by the unit norm direction of the perturbation for each sample, where an enhancement scale of 0 reflects the original dataset. b-d) Transferability of enhancement between the three models. All accuracies were evaluated with 5-fold cross-validation, with error bars showing standard deviation across 10 random seeds.
} 
\label{fig:model_based_enhance}
\end{figure}

Gradients were computed as the model coefficients in SVM and LR (linear models), while Pytorch's autograd feature was used for the FNN.
All gradients (\textit{i.e.}, $\nabla_xA$ in Algorithm \ref{alg:model-enhance}) were normalized to have a Frobenius norm of 1 and then multiplied by the corresponding enhancement scale in Figure \ref{fig:model_based_enhance}. 
After creating an ``enhanced dataset," we evaluated enhanced performance with nested k-fold cross-validation, as one would do if receiving this dataset without any knowledge of the enhancement, with a grid search in SVM and LR for the L2 regularization parameter $C$=\{1e-4,1e-3,1e-2,1e-1,1\} within each fold.
Due to computational restraints, we did not perform a hyperparameter search for the FNN but used the same hyperparameters described above.
Enhancement brought prediction performance from 50\% to 99\% in all three models (Fig \ref{fig:model_based_enhance}a), despite similar feature values between the original and enhanced datasets ($r=0.99$).

Furthermore, we considered whether the enhancement attacks could transfer between models.
After data were enhanced using Algorithm~\ref{alg:model-enhance} for each of the three models (Figure~\ref{fig:model_based_enhance}a), we re-trained the other two types of models to assess transferability.
For example, in Figure~\ref{fig:model_based_enhance}b, data are first enhanced using SVM and Algorithm~\ref{alg:model-enhance}; then, the performance of that enhanced dataset is evaluated using LR and FNN.
Overall, we found that the enhancement attacks transferred between each of the three models (Figure~\ref{fig:model_based_enhance}b-d).

\subsubsection{GOAL \#2: Enhancement of a particular method}
\label{section:goal2_experiment}
Since the enhancement attacks above transferred between models, we investigated how enhancement may be targeted to a specific model.
Although there are countless machine learning models, we selected three of the most common models for functional connectivity data to perform a case study: SVM, LR, and a FNN.
We demonstrated the hypothetical scenario in which one may perturb a dataset such that a FNN outperforms simpler methods like SVM and LR. 

\begin{figure}[t!]  
\centering
\includegraphics[scale=0.4]{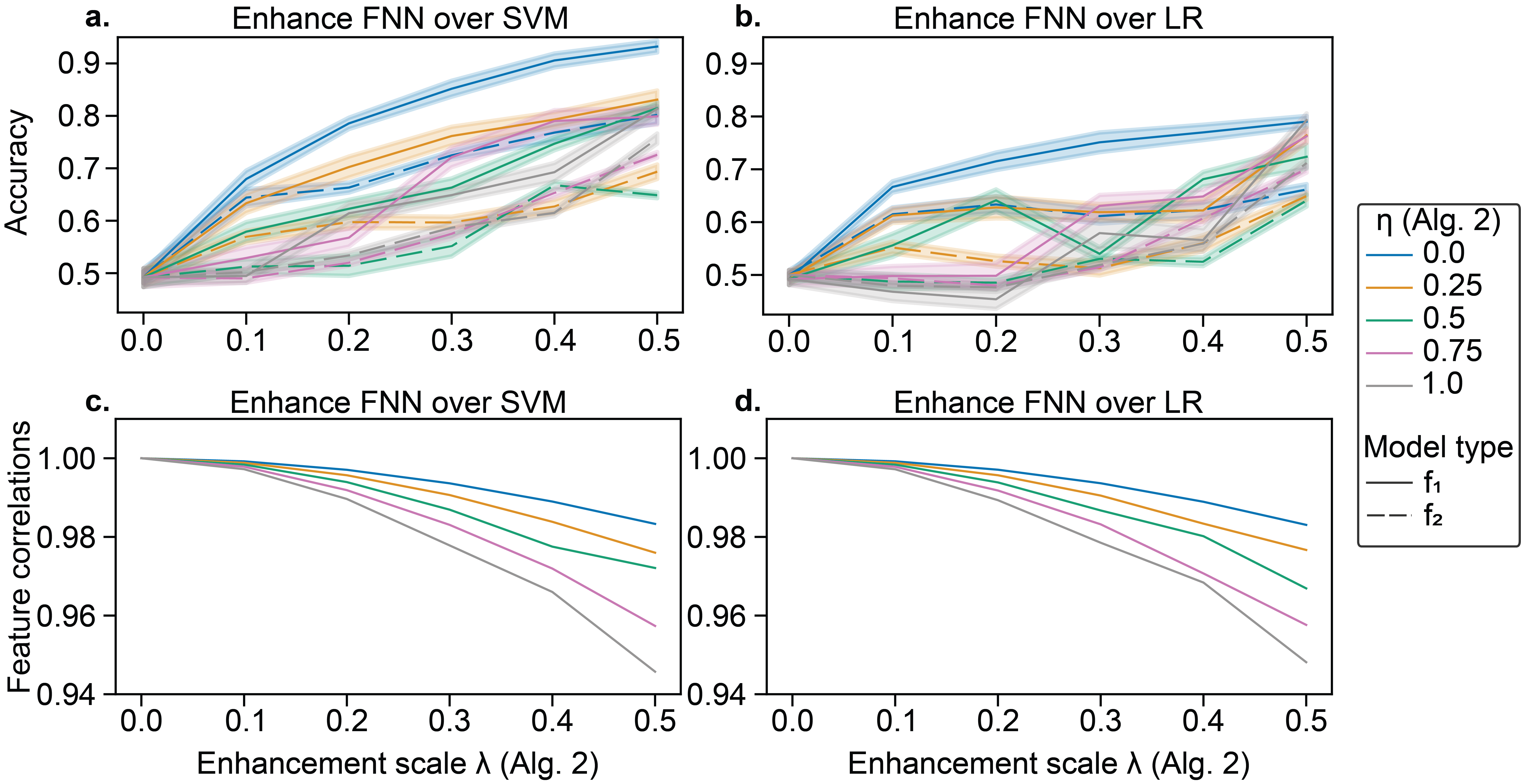}
\caption{
Enhancement of a FNN over a,c) SVM and b,d) LR in CNP. In a,b), data are enhanced with increasing $\lambda$ (see Algorithm \ref{alg:method-enhance}). Solid lines represent the accuracy for $f_1$ (FNN), while dashed lines show the accuracy for $f_2$ (SVM in a and LR in b). Error bars reflect standard deviation across 10 random seeds of k-fold cross-validation initialization seeds for FNN. Line color shows the suppression coefficient for $f_2$, $\eta$. In c,d), the correlation between original and enhanced features is shown with increasing $\lambda$. The original and enhanced features are highly correlated ($r's>0.94$).
} 
\label{fig:method_enhance}
\end{figure}

CNP data were manipulated following Algorithm~\ref{alg:method-enhance} to promote the performance of a particular method over another.
Hyperparameters were held constant when enhancing the data with Algorithm \ref{alg:method-enhance} (as detailed in GOAL \#1). 
After enhancing the data, we performed 10-fold cross-validation to evaluate accuracy with nested folds and a grid search for the L2 regularization parameter $C$ in SVM and LR.
We consider different enhancement scales $\lambda$ for the classifier of interest ($f_1$=FNN) and different suppression values $\eta$ for the classifier which we do not wish to perform well ($f_2$=SVM or LR).
Despite no differences in the original dataset, FNN outperformed SVM and LR (Fig \ref{fig:method_enhance}a-b), while maintaining high feature similarities (Fig \ref{fig:method_enhance}c-d).
The performance on $f_2$ generally increased, though less, as performance on $f_1$ increased, but increasing the suppression coefficient $\eta$ limited performance improvements of $f_2$.
Furthermore, attacks transferred between SVM and LR.
For $f_2$=SVM, $\lambda$=0.4, and $\eta$=0, accuracies for SVM and LR were 76.8\% and 77.8\% vs. 90.6\% for FNN.
For $f_2$=LR, $\lambda$=0.4, and $\eta$=0, accuracies for SVM and LR were 61.8\% and 62.3\%  vs. 77.0\% for FNN.
This method was less effective than that of Algorithm \ref{alg:model-enhance}, which is likely because the enhancement is weakened by the suppression coefficient and the projection of the gradients.

\subsubsection{Proof-of-concept validation in other models and datasets} 
While the focus of this work is enhancement of SVM, LR, and FNN models in the CNP dataset, we also wanted to demonstrate that other datasets and models were easily manipulated. To demonstrate generalizability to another dataset, we performed enhancement attacks with enhancement scale $\lambda=2$ in the Philadelphia Neurodevelopmental Cohort \cite{Satterthwaite2014-md,Satterthwaite2016-xm} (N=1126) dataset to predict self-reported sex using resting-state functional connectivity data. The accuracies were (baseline/after enhancement): SVM (79.40\%/100\%), logistic regression (79.13\%/100\%), and FNN (76.47\%/99.91\%). 

Furthermore, we demonstrated our results in the CNP dataset with another model, BrainNetCNN \cite{Kawahara2017-qa}, a deep learning model for brain connectivity data. BrainNetCNN consists of two edge-to-edge filters, one edge-to-node filter, one node-to-graph filter, and three dense layers. Further details of BrainNetCNN are in the original paper \cite{Kawahara2017-qa}. BrainNetCNN was trained with cross entropy loss and the Adam \cite{Kingma2014-cn} optimizer, with a learning rate of $\alpha=0.001$ and batch size of 10 for 20 epochs. For the gradient-based enhancement attack with enhancement scale $\lambda=3$, the resulting accuracy was 79.31\% (baseline: 41.10\%). We enhanced BrainNetCNN over the FNN with $\lambda=0.15$, $\eta=0$; accuracy was 85.43\% for BrainNetCNN and 52.33\% for FNN. We enhanced the FNN over BrainNetCNN with $\lambda=0.15$, $\eta=0$; accuracy was 80.16\% for BrainNetCNN and 99.27\% for FNN. These results demonstrate that complex models are also susceptible to designed data manipulations. 

\section{Discussion}
In this work, we developed a gradient-based framework for enhancement attacks and showed that a functional neuroimaging dataset could be modified to achieve essentially any desired prediction performance.
The vulnerability of machine learning pipelines to possible fraud (enhancement) is integral to the trustworthiness of the field. Considering the prevalence of research fraud \cite{Al-Marzouki2005-jb,Bik2016-ud,Piller2022-cf} and the increasing popularity of machine learning in biomedical research, we believe that enhancement could become a major issue, if it is not already.

\label{section:goal1_discussion}
For GOAL \#1, we demonstrated that a four-way classification task went from near-chance performance, where chance is defined as the most frequently occurring group (47.76\%), to over 99\% accuracy while the original and enhanced data remained highly similar ($r>0.99$). 
In a hypothetical scenario, a malicious researcher could collect data, enhance it to perform well in a specific classification or regression task, and release this data on a public repository.
For example, if the enhanced CNP dataset in this paper was released, the scientific community would be excited and impressed by the near-perfect classification accuracy between participants with no diagnosis, schizophrenia, bipolar disorder, and ADHD. 
In the academic sector, these false results could lead to hiring and grant decisions or the distribution of future grant funding under false pretenses. 
In the industrial sector, these false results may cause increased investments in certain companies.
Furthermore, enhancement attacks could have additional downstream effects, such as the undermining of public trust or the wasted time and resources of other researchers attempting to build upon the false results. 

\label{section:goal2_discussion}
For GOAL \#2, we found that in the best case, a FNN outperformed LR and SVM by 17\% in the enhanced dataset, despite no differences in original performance and high similarity between original and enhanced data ($r=0.99$).
The feasibility of modifying a dataset such that a specific type of model outperforms another is both powerful and potentially dangerous.
For instance, a start-up company may demonstrate that their new model outperforms the current industry standard to increase their valuation.
Alternatively, researchers may perturb a dataset so their novel method performs the best, leading to a (falsely) more exciting paper.   

There were several limitations to our study.
First, we investigated enhancement only in functional neuroimaging.
Future work should expand these concepts to other disciplines, which may have different sample sizes or dimensionality.
Second, the method enhancement algorithm (Algorithm \ref{alg:method-enhance}) only allows for the suppression of a specific model, and it remains to be seen how this can be extended to improve the performance of one model over many other models.
However, we demonstrated that the method-specific enhancement transferred across SVM and LR, which is promising for the extension of it to multiple models.
Third, we evaluated enhancement attacks in processed connectome data, and applying enhancement earlier in the processing pipeline (e.g., raw fMRI data) may reduce its effectiveness.
Fourth, future work should evaluate gradient-based enhancement of additional model types.
By design, the gradient-based enhancement attack is generalizeable and should be able to work for any machine learning model, including both classification and regression models, for which a gradient of the loss with respect to the input can be computed.

In conclusion, although our analysis was restricted to functional neuroimaging, these problems extend to the greater biomedical machine learning communities, where many view data and code sharing as the panacea for trustworthiness. 
Whereas adversarial attackers only have limited access to the model and data, enhancement attackers have unrestricted access, which makes developing defenses more difficult.
Still, future work should explore whether existing adversarial defenses, including augmentation and input transformations \cite{Ren2020-oc}, can be adapted to defend against enhancement attacks, though we expect these defenses will be less effective given the much greater capabilities of enhancement relative to adversarial attacks.
Another possible defense is data provenance tracking, such as DataLad \cite{Halchenko2021-zl}. 
However, one caveat is that data could be manipulated before provenance tracking begins.
Thus, an alternative solution could be the replication of studies in an independent dataset.
Overall, we hope that this work sparks additional discussion about possible defenses against enhancement attacks and data manipulations to secure the integrity of the field.

\subsubsection{Data use declaration and acknowledgment}
The UCLA Consortium for Neuropsychiatric Phenomics (download: \url{https://openneuro.org/datasets/ds000030/versions/00016}) and the Philadelphia Neurodevelopmental Cohort (dbGaP Study Accession: phs000607.v1.p1) are public datasets that obtained consent from participants and supervision from ethical review boards. We have local human research approval for using these datasets. 

%
%
%
%



\bibliographystyle{splncs04} 
\bibliography{refs} 

\end{document}

%% file: methods_javid.tex
In the following sections, we considered two separate attacker goals: falsely enhancing 1) overall classifier performance (GOAL \#1), and 2) the performance of one method over another (GOAL \#2) (Fig \ref{fig:summary}). 
Whereas traditional adversarial attackers may have complete (``white-box") or limited (``black-box" or ``gray-box") knowledge of the model and/or data \cite{Biggio2018-ka}, enhancement attackers have even greater knowledge than the ``white-box" setting.
Our methods can modify the entire dataset, which includes both training and test data. 
These enhancement attacker assumptions mimic a realistic setting, where a researcher could modify their data to falsely improve performance and then publicly release the data such that their results are computationally reproduced by others. 
We assess all methods with two main metrics: classification accuracy in the enhanced dataset and similarity between the original and enhanced datasets, as measured by Pearson's correlation. 
A highly effective enhancement attack is one that falsely increases the accuracy while the original and enhanced datasets remain highly similar.

\subsubsection{GOAL \#1: Gradient-based enhancement}
\label{section:goal1}


The key idea behind gradient-based enhancement is to ``push" the samples in the direction of a learned model to make the decision boundaries clearer and more consistent across all samples, thus improving performance.
The method is similar to an adversarial evasion attack, except we are altering the entire dataset, not just the test data.
For a single held-out point, one may optimally change the classification by perturbing the point in the direction of $g=\nabla_xA$, where $A$ can be a decision function or loss function.
For example, in the case of binary linear support vector machine (SVM) or logistic regression (LR):
\begin{equation}
X_{held-out, y=-1} \gets X_{held-out, y=-1} - \epsilon*w
\label{eq:adv_grad_neg}
\end{equation}
\begin{equation}
X_{held-out, y=1} \gets X_{held-out, y=1} + \epsilon*w
\label{eq:adv_grad_pos}
\end{equation}
where $w$ is a vector of model coefficients and $\epsilon$ is a scaling factor.
Equations~\ref{eq:adv_grad_neg}-\ref{eq:adv_grad_pos} would move the corresponding held-out points toward the correct side of the decision boundary. 
As summarized in Algorithm~\ref{alg.model-based}, a model $f$ is first trained by holding one or numerous points out with K-fold partitioning. 
Then, the held-out point(s) are updated with the attacker gradient $g=\nabla_xA$, such that the model will predict them correctly. This process repeats until all points in the dataset are held out. 
Empirically, we found that enhancement was most effective for complex models (i.e., neural networks) when updating the held-out data iteratively within the cross-validation loop, whereas updating the dataset once after the cross-validation loop was more effective for simpler models (i.e., support vector machine or logistic regression).
Since learned model coefficients are similar when only holding out a small fraction of the points, this method  pushes all points of a given class in a consistent direction. Eventually, when the enhanced dataset is released, an independent researcher would not notice any perturbations but would falsely find higher performance.  

\begin{algorithm}[t!]
\centering
\caption{Gradient-based enhancement attacks}
\label{alg:model-enhance}
\begin{algorithmic}
\State $D \in\{X, y\}$: dataset
\State $f$: model
\State $n_{folds}$: folds for K-fold partitioning
\State $\lambda$: enhancement step size
\For{$k = 1 : n_{folds}$}
    \State Establish $D_{tr}, D_{held-out}$
    \State Train $f$
    \State $g \gets \nabla_{x}A$ where $A=L(f, x)$ or $DF(f, x)$ 
    \State $X_{held-out} \gets X_{held-out}  - \lambda g$ 
\EndFor
\end{algorithmic}
\label{alg.model-based}
\end{algorithm}

\subsubsection{GOAL \#2: Enhancement of a particular method}
\label{section:goal2}

A roadblock to method-specific enhancement is that the gradients used in Equations~\ref{eq:adv_grad_neg}-\ref{eq:adv_grad_pos} generally transfer well across model types \cite{Demontis2019-oy}, which would make this process ineffective in enhancing the performance of a specific method over another. 
Transferability of attacks from a base classifier $f_1$ to another classifier $f_2$ is defined by \cite{Demontis2019-oy} as how well an attack designed for $f_1$ works on $f_2$.  
In this case, we do \textit{not} wish for the attacks to transfer between models. We want to find a new direction $g_1^\prime$ that enhances the performance of $f_1$ but does not affect $f_2$.
We achieve this by taking the component of $g_1$ that is orthogonal to $g_2$:
\begin{equation}
g_1^\prime = proj_{g_{2}^{\bot}}(g_1)
\label{eq:model_based_proj}
\end{equation}
Furthermore, Equation \ref{eq:model_based_proj} may not be sufficient to limit the performance of $f_2$, since $f_2$ can learn a new decision boundary after retraining. As such, we propose to include a term $g_2^\prime$ to suppress the performance of $f_2$:
\begin{equation}
g_2^\prime = proj_{g_{1}^{\bot}}(g_2)
\label{eq:suppression_proj}
\end{equation}
Then, for a held-out sample, we can update it as follows to attempt to improve the performance of $f_1$ but not $f_2$:
\begin{equation}
    x^\prime = x - \lambda(g_1^\prime - \eta g_2^\prime)
\label{eq:model_based_update}
\end{equation}
where $\lambda$ and the suppression coefficient $\eta$ control the influence of $g_1^\prime$ and $g_2^\prime$.

Similar to the model-based data enhancement, we split the data into $k$ folds. For each partitioning, we train two models: 1) A model that we want to enhance (i.e., $f_1$), and 2) a second model that we do not want to enhance (i.e., $f_2$). 
Subsequently, Equations \ref{eq:model_based_proj}-\ref{eq:model_based_update} are applied to update the held-out data, and the process is repeated until each sample is held out once.

\begin{algorithm}[t!]
\centering
\caption{Method enhancement}\label{alg:method-enhance}
\begin{algorithmic}
\State $D \in\{X, y\}$: dataset
\State $f_1$: model to enhance
\State $f_2$: model to avoid enhancement
\State $n_{folds}$: number of folds for K-fold partitioning
\State $\lambda$: enhancement step size
\State $\eta$: enhancement suppression coefficient
\For{$k=1 : n_{folds}$}
    \State Establish $D_{train}, D_{held-out}$
    \State Train $f_1, f_2$
    \State $g_1 \gets \nabla_{x}A(f_1, X_{held-out})$
    \State $g_2 \gets \nabla_{x}A(f_2, X_{held-out})$
    
    \State $X_{held-out} \gets X_{held-out}   - \lambda ( proj_{g_{2}^{\bot}}(g_1) - \eta \ proj_{g_{1}^{\bot}}(g_2) )$
\EndFor
\end{algorithmic}
\end{algorithm}